\documentclass[conference]{IEEEtran}
\usepackage{graphicx}
\graphicspath{{./figures/}}
\begin{document}

\title{Leveraging Product as an Activation Function in Deep Networks}
\author{
    \IEEEauthorblockN{Luke B. Godfrey}
    \IEEEauthorblockA{
        SupplyPike\\
        Fayetteville, AR 72703\\
        Email: luke@supplypike.com}
    \and
    \IEEEauthorblockN{Michael S. Gashler}
    \IEEEauthorblockA{
        University of Arkansas\\
        Department of Computer Science and Computer Engineering\\
        Fayetteville, AR 72712\\
        Email: mgashler@uark.edu}}
\maketitle

\begin{abstract}
Product unit neural networks (PUNNs) are powerful representational models with a strong theoretical basis, but have proven to be difficult to train with gradient-based optimizers.
We present windowed product unit neural networks (WPUNNs), a simple method of leveraging product as a nonlinearity in a neural network.
Windowing the product tames the complex gradient surface and enables WPUNNs to learn effectively, solving the problems faced by PUNNs.
WPUNNs use product layers between traditional sum layers, capturing the representational power of product units and using the product itself as a nonlinearity.
We find the result that this method works as well as traditional nonlinearities like ReLU on the MNIST dataset.
We demonstrate that WPUNNs can also generalize gated units in recurrent neural networks, yielding results comparable to LSTM networks.
\end{abstract}

\IEEEpeerreviewmaketitle

\section{Introduction}

Nodes in an artificial neural network traditionally apply a nonlinearity to a weighted sum of its inputs.
Previous work suggests that using a product rather than a sum should increase the representational power of a neural network, but these product unit neural networks (PUNNs) have proven difficult to train using gradient-based optimizers \cite{Engelbrecht99trainingproduct}.
Given an input of $N$ elements, a product unit multiplies together all $N$ elements raised to arbitrary powers (where each exponent is a learned parameter).
It is not surprising, then, that gradient descent has difficulty training networks of product units; the derivative of an $N$ element product with respect to one of its elements is an $N-1$ element product.
Thus, the error surface of a PUNN is particularly chaotic and contains many poor local optima.

We present windowed product unit neural networks (WPUNNs), a simple method of leveraging product as a nonlinearity in a neural network.
Windowing the product tames the complex gradient surface and enables WPUNNs to learn effectively, solving the problems faced by PUNNs.
A windowed product unit takes the product not of all $N$ inputs, but on a small portion of the inputs (a window), significantly reducing gradient complexity.
WPUNNs use layers of these windowed product units between traditional sum layers, capturing the representational power of product units and using the product itself as a nonlinearity.
In this paper, we make three discoveries related to WPUNNs:

\begin{itemize}
    \item gradient-based optimization can be used effectively with windowed units,
    \item windowed product is as effective as traditional nonlinearities like rectified linear units (ReLU), and
    \item WPUNNs can generalize gated units in recurrent neural networks.
\end{itemize}

The rest of this paper is laid out as follows.
Related work is briefly discussed in Section~\ref{sec:related}.
In Section~\ref{sec:approach}, we present the formulation of WPUNNs.
Section~\ref{sec:applications} lays out a couple of theoretical applications of WPUNNs.
We present our findings in Section~\ref{sec:results} and offer some concluding thoughts in Section~\ref{sec:conclusion}.

\section{Related Work}
\label{sec:related}

\subsection{Product Unit Neural Networks}

Product unit neural networks (PUNNs) were introduced in the late 1980s as computationally powerful alternatives to traditional summation unit networks \cite{durbin1989product}.
In a PUNN, weights are used as exponents rather than as coefficients, giving a single product layer followed by a sum enough representational power to represent arbitrary polynomials.
The standard forumulation of a PUNN unit is

\begin{equation}
    y = \Pi_{i=1}^N x_i^{\theta_i},
\end{equation}

where $y$ is the output, $N$ is the number of inputs, $x_i$ is the $ith$ input, and $\theta_i$ is the $ith$ weight.
The use of product instead of sum increases a network's information capacity \cite{janson1993training, wang2000internal}
and enables higher-order combinations of inputs \cite{Engelbrecht99trainingproduct}.
The two drawbacks to this approach are the overhead of computing so many exponents and logarithms \cite{janson1993training} and the difficulty of training a PUNN with gradient-based optimization methods \cite{Engelbrecht99trainingproduct}.
This second problem has proven to be particularly challenging, so most of the work on PUNNs focus on alternative,
non-gradient-based techniques such as genetic and evolutionary algorithms \cite{janson1993training, Engelbrecht99trainingproduct, ismail2000global, hervas2006classification, hervas2007logistic, martinez2008evolutionary}
and particle swarm optimizers \cite{Engelbrecht99trainingproduct, ismail2000global, van2001training}.
Despite the difficulty in training them, PUNNs have been applied to a number of problems, including
classification of Listeria growth \cite{martinez2008evolutionary}, massive missing data reconstruction \cite{duran2016massive}, and time series forecasting \cite{fernandez2016time}.
More recent work has focused on hybridizing PUNNs with sigmoid and RBF networks \cite{gutierrez2008logistic, gutierrez2009combined, gutierrez2010hybridizing, redel2013ensembles},
and developing improved evolutionary algorithms for training PUNNs \cite{tallon2011improving, tallon2013feature, guerrero2016experimental}.
Our work discovers a way to capture the representational power of a PUNN without losing the ability to train the network using standard gradient-based optimization algorithms.

\subsection{Gated Units}

Recurrent neural networks often make use of a product to implement gates that control the flow of information in a sequence.
Long short term memory (LSTM) \cite{hochreiter1997long,gers1999learning} and gated recurrent unit (GRU) \cite{cho2014learning,chung2014empirical} networks use a combination of sigmoids and products to ``gate'' information.
Gated units use the input and recurrent connections to compute several gate values (in the range $[0..1]$) and a memory value.
Some gated units use one of the gate values as an input gate, controlling how much of the input is used when calculating the new memory value.
Some have a forget gate, controlling how much of the previous memory value is used when calculating the new memory value.
Still others have an output gate, controlling how much of the memory value is actually used as output.
Gates and outputs are generally computed using a standard feedforward approach (a nonlinearity applied to a weighted sum of inputs), then multiplied in a pre-defined way depending on which model is used (for example, LSTM or GRU).
LSTM networks, in particular, have proven to be particularly powerful models for
speech recognition \cite{graves2013hybrid},
language modeling \cite{gers2001lstm},
text-to-speech synthesis \cite{fan2014tts}, and
handwriting recognition and generation \cite{graves2009offline,graves2013generating}.
Our approach is related to gated units by the use of multiplication and can be used as a generalization of gated units.

% \begin{figure*}
%     \label{fig:vary_s}
%     \includegraphics[width=\textwidth]{vary_s.png}
%     \caption{Test set misclassifications of a WPUNN on the MNIST dataset, fixing $w = 4$ but varying $s$. This supports our theory that WPUNNs are not sensitive to the hyperparameter $s$ (stride length).}
% \end{figure*}

\subsection{Sum-Product Networks}

The use of the product operation has been shown to be effective in graphical inference models.
In 2011, Poon et. al proposed a deep architecture for graphical inference models called Sum-Product Networks (SPN).
This architecture not only has a strong mathematical basis, but also yields compelling results, surpassing state-of-the-art deep neural networks at the image completion task \cite{poon2011sum}.
A SPN is arranged into layers, as in a multilayer perceptron.
Inputs to a SPN are binary (either $0$ or $1$), and layers alternate between weighted sums and weighted products (where the weights used in sums are in $[0..1]$ and the weights used in products are either $0$ or $1$).
SPNs have been successfully used for
classifying videos of various activities \cite{amer2012sum,amer2016sum},
bandwidth extension of speech signals \cite{peharz2014modeling},
image classification for autonomous flight \cite{sguerra2016image},
and more \cite{li2015combining,krakovna2016minimalistic,melibari2016dynamic}.
Our work borrows the idea of alternating sum and product as used in SPN graphical models.
Although neural networks have much in common with graphical models, they are better-suited for learning in domains where little is known {\em a priori} about the relations among available attributes.

\section{Approach}
\label{sec:approach}

In our approach, we use a specialized kind of product unit in a neural network.
Our method has two key differences from the standard product unit used in PUNNs.
First, our product is weightless; we do not use weights as exponents in the product.
Second, our product is windowed; rather than having each unit take a product of all elements in the input vector, each unit takes a product of a small window of the inputs.
WPUNNs have two hyperparameters: the window size $w$, $1 \leq w \leq N$, and stride length $s$, $1 \leq s \leq w$.
Given a vector of input values, we slide a window of size $w$ using a stride of $s$ and multiply all elements in the window to yield an output value.
In the simplest case, where $w = 2$ and $s = 2$, every odd input is multiplied by its adjacent even input and the input vector is reduced by a factor of $2$.
Formally, for an input vector $x$ of size $N$, the output is a vector $y$ of size $M = (N - w + s - 1) / s + 1$ as defined by the following equation, for $1 \leq i \leq M$:

\begin{equation}
    \label{eq:pp}
    y_i = \Pi_{j=0}^{w-1}{x_{si+j}}.
\end{equation}

Note that changing the product to the $max$ aggregation function changes the operation into max pooling, which is commonly used in convolutional neural networks \cite{krizhevsky2012imagenet}.

By inserting a layer of these windowed product units between each fully-connected layer in a neural network, we build a windowed product unit neural network (WPUNN).
We make the following observations about a WPUNN unit:

\begin{itemize}
    \item It is differentiable; $\frac{dy_i}{dx_j} = y_i / x_j$.
    \item It is a generalization of PUNN units with respect to window size; a WPUNN unit with $w = N$ is equivalent to a PUNN unit.
    \item It is a specialization of PUNN units with respect to weights; a PUNN unit where all weights are $1$ is a WPUNN unit.
    \item Its derivative is less chaotic than the derivative of a PUNN unit; all exponents are 1, and (generally) $w < N$.
    \item It is nonlinear with respect to each input, because it is multiplied by a variable and not a scalar.
\end{itemize}

These observations yield a number of corollaries.
First, because WPUNN derivitives are less chaotic than PUNN derivatives, WPUNNs are able to learn effectively without any other explicit nonlinearity (such as rectified linear or sigmoidal units).
With a small enough window size, such a network should not experience the same training problems as experienced in training PUNNs, which should enable it to learn using standard gradient-based optimization techniques rather than resorting to evolutionary algorithms.
Second, because WPUNN layers are weightless, the forward and backward propagation steps are more efficient than in PUNNs.
Third, because a WPUNN with a high $w$ is more like a PUNN, WPUNNs are sensitive to this hyperparameter and will favor lower values of $w$.
Fourth, because a fully-connected layer preceeds all WPUNN units, WPUNNs are not sensitive to the stride length hyperparameter $s$.
With a single fully-connected layer, a neural network can weight, rearrange, and duplicate inputs.
Thus, a WPUNN layer after a sufficiently wide fully-connected layer does not need stride to overlap window, and a WPUNN should be as effective with $s = 1$ as with $s = w$.

\begin{figure*}
    \label{fig:vary_w}
    \includegraphics[width=\textwidth]{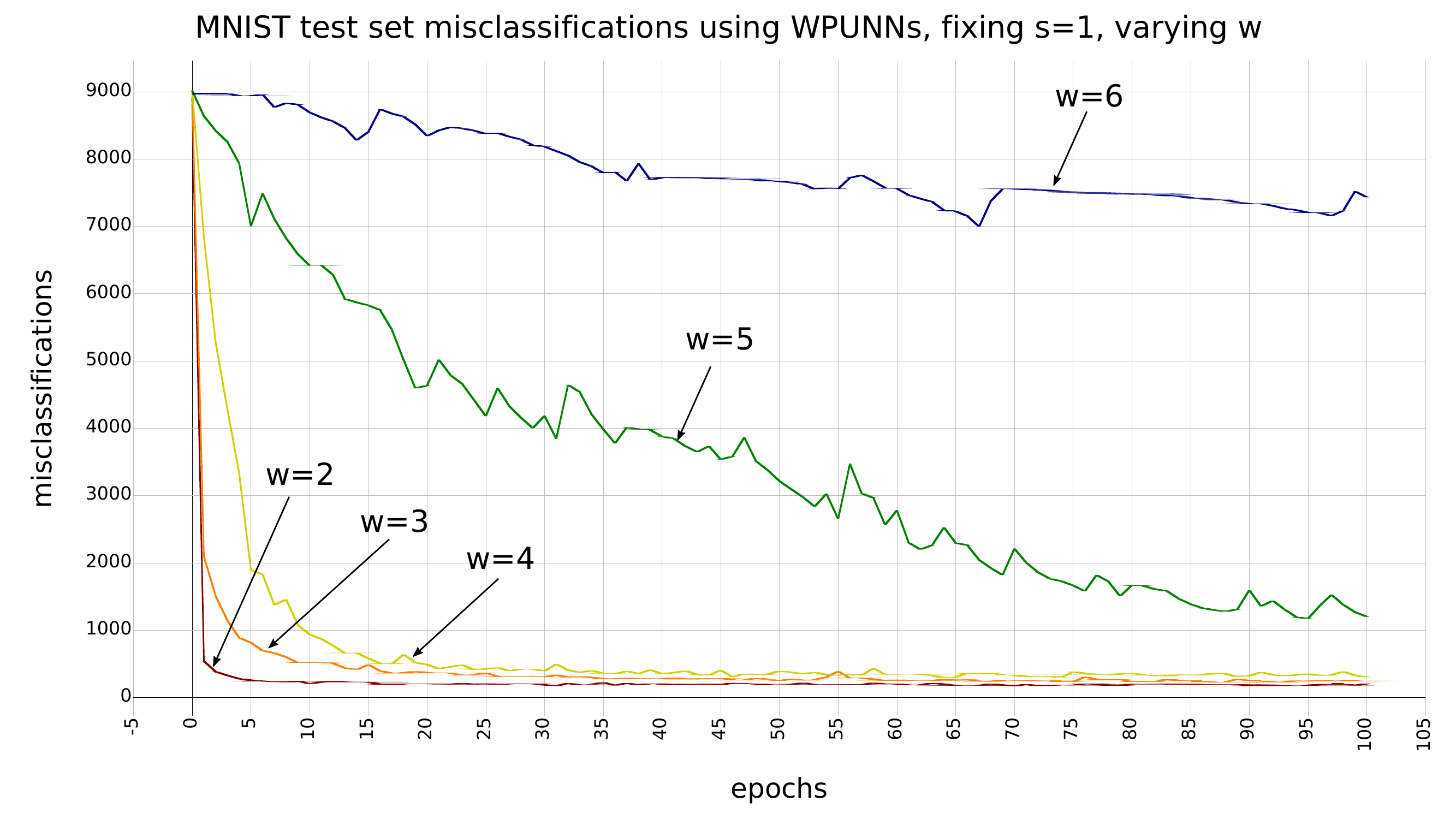}
    \caption{Test set misclassifications over time of a WPUNN on the MNIST dataset, fixing $s = 1$ but varying $w$. The inverse relationship between accuracy and window size demonstrates that WPUNNs are sensitive to the hyperparameter $w$ (window size). Reducing window size solves the major problems associated with training PUNNs, which validates our primary contribution.}
\end{figure*}

\section{Applications}
\label{sec:applications}

\subsection{Representing Polynomials}

A WPUNN can exactly represent arbitrary polynomials.
Although this appears to be a simple problem, neural networks with standard nonlinearities such as $tanh$ and $ReLU$ can only, at best, approximate these values.
Representing polynomials is particularly useful when modeling dynamical systems, which are often derived from polynomial equations \cite{jarrah2007reverse}.

WPUNNs can model any monomial, $x^d$, as a network with a one input and one hidden layer.
The input is simply $x$.
The hidden layer is a fully-connected layer of width $d$, with all weights set to $1$ and all biases set to $0$.
This effectively duplicates the input $d$ times.
Finally, the output layer is a WPUNN layer with window size $w = d$ (i.e. a single product).
To extend this to polynomials, we can design a network that repeats this construction for each monomial in the polynomial, using a window size corresponding to the monomial with the highest degree and a stride of the same size.

WPUNNs are not unique in their ability to represent polynomials.
In fact, PUNNs can exactly represent arbitrary polynomials in a single {\em unit} rather than in two layers, as weights in PUNNs are exponents.
However, PUNNs have proven to be difficult to train with traditional gradient-based techniques, while WPUNNs do not have this problem.

\subsection{Generalizing Gated Units}
\label{subsec:gated}

WPUNNs can generalize gated units.
Gated units depend on the multiplication of inputs, outputs, and hidden states (``cell value'' in LSTM) with their corresponding gate values.
Because these values are computed as dot products, the gating operation is equivalent to a WPUNN layer with $w = s = 2$.
Thus, we can design a WPUNN-based gated unit architecture that similarly controls the flow of information.

One possible architecture for gated units using WPUNNs is the following 3-layer block, where $N$ is the number of outputs.
The first layer is a fully-connected layer that takes $x_t$ (the current input) and $y_{t-1}$ (the output at the previous time step) as inputs and produces $2N$ values.
This allows the network to rearrange the inputs as needed; if we assume that existing RNN units are optimal, this will interleave elements of $x_t$ with ``input gate values'' and the elements of $y_{t-1}$ with ``output gate values''.
The second layer is a sigmoid that flattens the output from the first layer into the range $[0..1]$.
Rather than using one nonlinearity for squashing the input and output and another for squashing the gates, we use the same nonlinearity for both.
The final layer is a windowed product layer with $w = s = 2$.
This layer effectively gates the odd-indexed units by the even-indexed units (or vice versa) and reduces the $2N$ values from the first two layers to $N$ values, which is the target number of outputs.
We present a comparison of this proposed architecture with long short term memory (LSTM) networks in Section~\ref{sec:results}.

\begin{figure*}
    \label{fig:poly}
    \includegraphics[width=\textwidth]{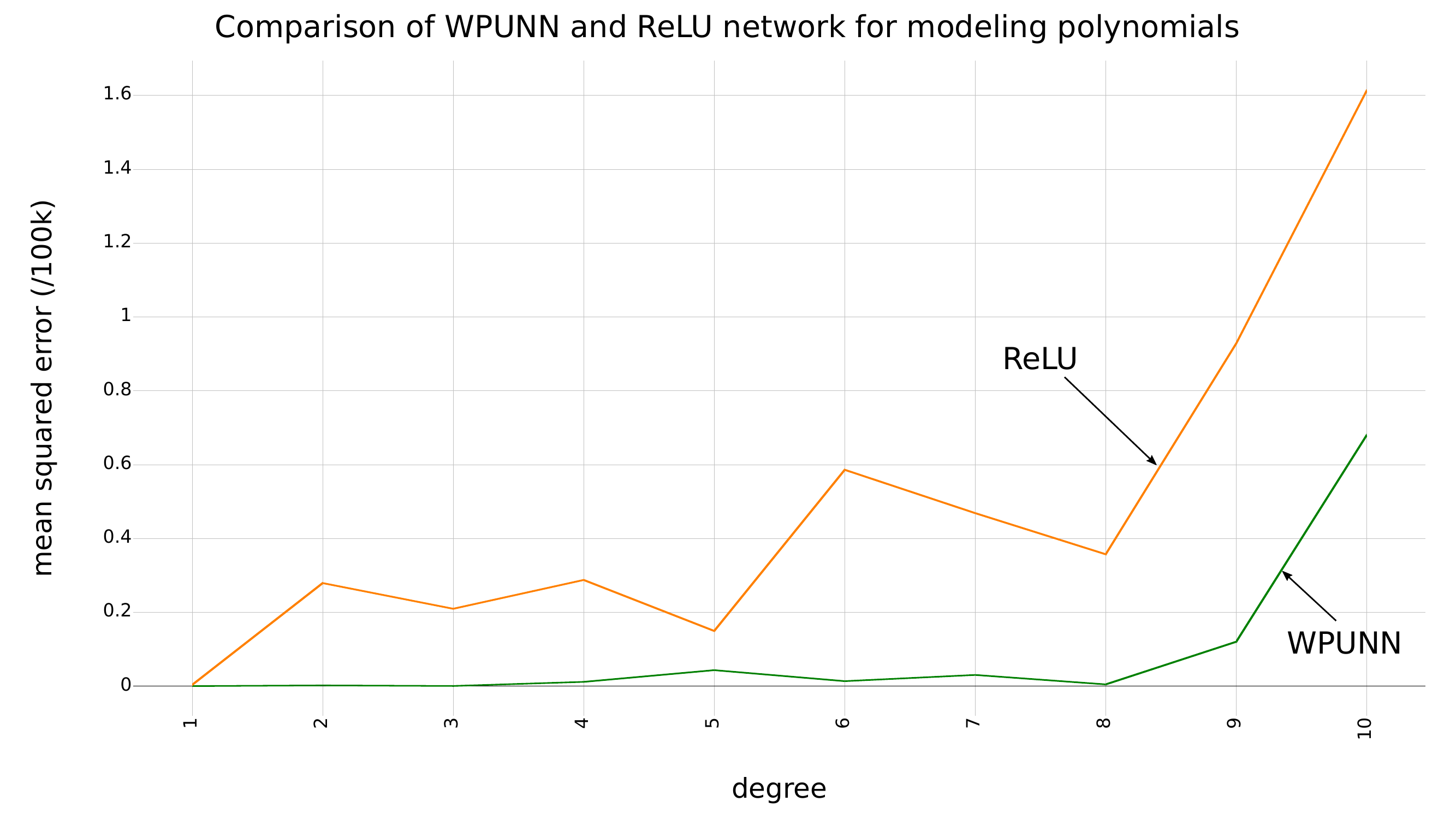}
    \caption{A comparison of two neural networks for modeling polynomials. The vertical axis is error (lower is better) and the horizontal axis is polynomial degree. The green curve is the error rate from a WPUNN and the orange curve is the error rate from a ReLU network. As expected, the WPUNN consistently yields a lower error rate than the ReLU model. The noticable increase in loss in the WPUNN for $d = 9$ and $d = 10$ can be attributed to the depth of the network being less than $log(d)$.}
\end{figure*}

\section{Results}
\label{sec:results}

In our first two tests, we use the MNIST dataset to evaluate our claims that WPUNNs are insensitive to the hyperparameter $s$ (stride length) and that they are sensitive to the hyperparameter $w$ (window size).
In both tests, we use a WPUNN with the following 6-layer topology:
1) a fully-connected layer to 300,
2) a WPUNN layer (parameters $s$ and $w$ vary),
3) a fully-connected layer to 100,
4) a WPUNN layer (parameters $s$ and $w$ vary),
5) a fully-connected layer to 10, and
6) a log soft max layer.
The network is trained to minimize negative log loss using the Adam optimizer with a learning rate of $1e^{-4}$.

For the first experiment, to test our claim that WPUNNs are insensitive to the hyperparameter $s$ (stride length), we try varing the hyperparameters while training on MNIST.
For this test, we fix $w = 4$ and vary $s$ from 1 to 4.
If WPUNNs are not sensitive to the choice of $s$, we would expect to see a similar learning curve for these four models.
All four models we tested performed with near-identical success, with an average test set misclassification rate of 3.56\% and a variance of 0.29.
This supports our theory that WPUNNs are not sensitive to the hyperparameter $s$.

For the second experiment, to test our claim that WPUNNs are sensitive to the hyperparameter $w$ (window size), we fix $s = 1$ and vary $w$ from 2 to 8.
If WPUNNs are sensitive to $w$, we would expect to see a higher learning accuracy on the models near $w = 2$ and a lower learning accuracy on the models near $w = 8$.
Figure~\ref{fig:vary_w} shows a plot of the test set error over time for these models.
As expected, there is an inverse correlation between $w$ and the accuracy, degrading dramatically when $w > 4$.
This supports our theory that WPUNNs are sensitive to $w$, and $w = 2$ yields higher accuracy.
Thus, reducing window size solves the major problems associated with training PUNNs, which validates our primary contribution.
As explained in Section~\ref{sec:approach}, fixing $w = s = 2$ does not significantly affect the representational power of a WPUNN that is sufficiently deep (logarithmic with respect to degree) and wide (at most quadratic with respect to the number of inputs).

\begin{figure*}
    \label{fig:co2}
    \includegraphics[width=\textwidth]{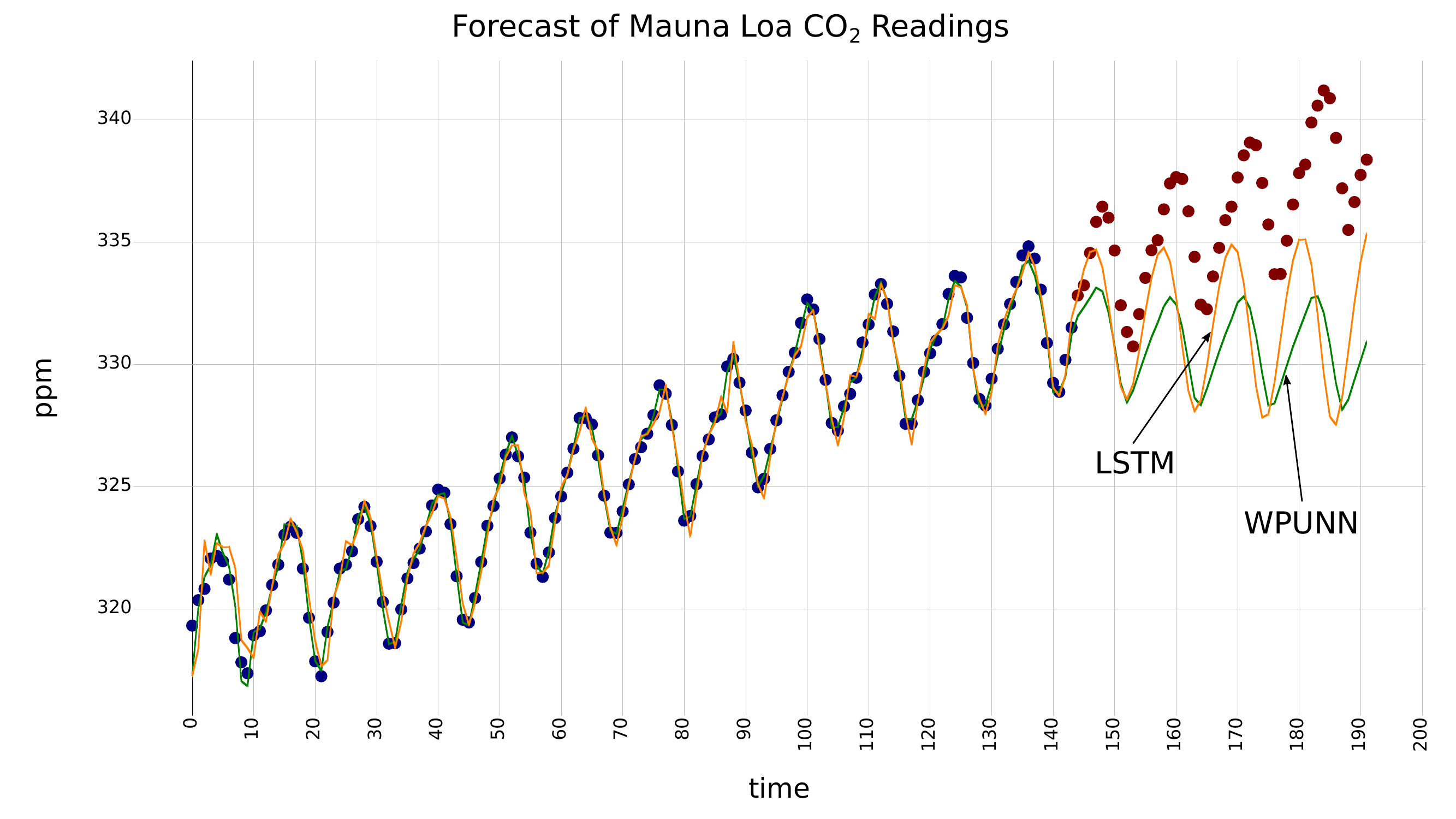}
    \caption{Forecasts made by two models on a time-series of Mauna Loa CO2 readings. Blue points represent training data, red points represent withheld testing data, the green curve represents the forecast by a WPUNN, and the orange curve represents the forecast by a LSTM network. Although the LSTM network yields a more accurate prediction, the WPUNN model is faster and uses a simpler architecture.}
\end{figure*}

In our third experiment, we test how well a WPUNN can represent polynomials by generating a random polynomial, sampling training data from that polynomial, then testing how accurately a trained WPUNN can calculate the polynomial value given a new set of values for the variables.
We generate a polynomial with two variables, $x$ and $y$, and one term for each monomial of degree $d$ or less (varying $d$ in our experiments).
The coefficients on the terms are drawn from a uniform distribution in $[-1, 1]$.
We then generate 1000 training samples by selecting values for $x$ and $y$ also drawn from a uniform distribution in $[-1, 1]$, using the values of $x$ and $y$ as features and the computed values of the polynomial as labels.
We repeat this generation to obtain 1000 testing samples that are withheld from the model.
We compare two models with similar topologies.
The first model is a WPUNN with the following topology:
1) a fully-connected layer to 50,
2) a WPUNN layer,
3) a fully-connected layer to 50,
4) a WPUNN layer,
5) a fully-connected layer to 50,
6) a WPUNN layer, and
7) a fully-connected layer to 1.
The second model is identical except that, instead of WPUNN, we use a leaky ReLU activation with parameter $\ell = 0.1$.
With this construction, the WPUNN model has 2776 parameters while the ReLU model has 5301 parameters, and the WPUNN trains approximately twice as fast as the ReLU network.
The networks are trained to minimize mean squared error using the Adam optimizer with a learning rate of $1e^{-3}$.
We repeat the experiment varying the degree of the polynomial from $d = 1$ to $d = 10$ and plot to results in Figure~\ref{fig:poly}.
As expected, the WPUNN consistently yields a lower error rate than the ReLU model.
The noticable increase in loss in the WPUNN for $d = 9$ and $d = 10$ can be attributed to the depth of the network being less than $log(d)$.

% WPUNN params for the time-series results:
% epochs              = 5000
% learningRate        = 0.01
% seed                = 0
% sequenceLength      = 36

% WPUNN-no-f error: 0.0153394
% LSTM-no-f error:  0.0058001
% WPUNN-f error:    0.122066
% LSTM-f error:     0.0851832

In our fourth experiment, we test how well the gated unit construction proposed in Section~\ref{subsec:gated} works.
We compare a WPUNN with a LSTM network on the Mauna Loa CO2 ppm time series \cite{hipel1994time}.
The WPUNN uses the following topology:
1) a fully-connected layer to 100,
2) a sigmoid activation,
3) a WPUNN layer with a recurrent connection to layer 1,
4) a fully-connected layer to 100,
5) a sigmoid activation,
6) a WPUNN layer with a recurrent connection to layer 4,
7) a fully-connected layer to 1.
The LSTM network uses the following topology:
1) a LSTM layer to 100,
2) a LSTM layer to 100,
3) a fully-connected layer to 1.
Despite the difference in depth (7-layer WPUNN as opposed to a 3-layer LSTM network), the WPUNN has 9026 parameters while the LSTM network has 27401 parameters, and the WPUNN trains approximately three times as fast as the LSTM network.
We withhold the last 25\% of the CO2 time-series data for testing and train both models on the first 75\%.
The networks are trained to minimize mean squared error using the Adam optimizer with a learning rate of $1e^{-2}$.
A plot of the forecasts from each model is shown in Figure~\ref{fig:co2}.
The blue points represent the training data.
The red points represent the withheld testing data.
The green curve is the forecast from the WPUNN, and the orange curve is the forecast from the LSTM network.
The mean squared error for the WPUNN forecast is 0.12, while the mean squared error for the LSTM forecast is 0.085.
The LSTM network outperforms the WPUNN model, but the WPUNN model performs notably well despite using only a third of the weights and a significantly simpler architecture.
It is worth noting that because WPUNN generalizes LSTM, if we used a more complex WPUNN architecture, the WPUNN predictions would be as accurate as the LSTM predictions.

\section{Conclusion}
\label{sec:conclusion}

The results presented in Section~\ref{sec:results} demonstrate that WPUNNs are effective machine learning models.
We make two primary contributions:
1) we give a construction of PUNNs that can be trained by gradient-based optimizers, and
2) we provide empirical results demonstrating the utility of using product as a nonlinearity in a neural network.
Without any other nonlinearity, WPUNNs work as well as traditional neural networks using fewer weights.
WPUNNs can be used to construct gated units in recurrent neural networks to model time-series data as effectively as LSTM networks.

%\IEEEtriggeratref{18}
\bibliographystyle{IEEEtran}
\bibliography{refs}

\end{document}